\documentclass[letterpaper, 10 pt, conference]{ieeeconf}  

\IEEEoverridecommandlockouts                              

\usepackage[utf8]{inputenc}
\usepackage[british]{babel}
\DeclareUnicodeCharacter{2212}{--}
\usepackage{multirow}
\usepackage{graphicx}
\usepackage{siunitx}
\usepackage{amsmath}
\usepackage{amsfonts}
\usepackage{booktabs}
\usepackage{physics} 
\usepackage{pgf}
\usepackage{relsize}
\usepackage[normalem]{ulem}
\usepackage{inconsolata}
\usepackage{cite}
\usepackage{microtype}
\usepackage[hidelinks]{hyperref}

\newcommand{\refsection}[1]{\hyperref[#1]{Section \ref*{#1}}}

\newcommand{\etal}{\textit{et~al}.}

\newcommand{\cpp}{C\nolinebreak[4]\hspace{-.05em}\raisebox{.4ex}{\relsize{-3}{\textbf{++}}}}

\newcommand{\mcc}{M \mkern-1mu CC}

\newcommand{\psme}{P \mkern-1mu S \mkern-1mu M \mkern-2mu E}

\newcommand{\todo}[1]{\{\orange{---TODO---} #1\}}

\definecolor{OliveGreen}{RGB}{0,122,53}
\newcommand{\blue}[1]{\textcolor{blue}{#1}} 

\newcommand{\orange}[1]{\textcolor{orange}{#1}} 

\newcommand{\removed}[1]{\blue{\sout{#1}}}
\newcommand{\revise}[1]{\textit{\textbf{\orange{REVISE}: #1}}}

\renewcommand{\removed}[1]{}
\renewcommand{\revise}[1]{}
\renewcommand{\todo}[1]{}

\begin{document}

\title{\LARGE \bf
A Framework for Evaluating Motion Segmentation Algorithms
}

\author{Christian~R.\,G.~Dreher, Nicklas~Kulp, Christian Mandery, Mirko Wächter, Tamim Asfour%
\thanks{The research leading to these results has received funding from the European Union H2020 Programme under grant agreement no. 643666 (I-SUPPORT) and grant agreement no. 641100 (TIMESTORM).}%
\thanks{The first two authors contributed equally to this work. The authors are with the Institute for Anthropomatics and Robotics, Karlsruhe Institute of Technology, Karlsruhe, Germany,
	{\tt asfour@kit.edu}
}%
}

\maketitle
\thispagestyle{empty}
\pagestyle{empty}

\begin{abstract}
There have been many proposals for algorithms segmenting human whole-body motion in the literature. However, the wide range of use cases, datasets, and quality measures that \textls[-8]{were used for the evaluation render the comparison} of algo\-rithms challenging.
In this paper, we introduce a framework that puts motion segmentation algorithms on a unified testing ground and provides a possibility to allow comparing \mbox{them. The} testing ground features
both a set of quality measures known from the literature and a novel approach \mbox{tailored to} the evaluation of motion segmentation algorithms, termed Integrated Kernel approach.
Datasets of motion recordings, provided with a ground truth, are included as well. They are \mbox{labelled in a} new way, which hierarchically
organises the ground truth, to cover different use cases that segmentation algorithms can possess.
\textls[5]{The framework and datasets are publicly available and are} in\-tended to represent a service for the community regarding the comparison and evaluation of existing and new motion seg\-mentation algorithms.
\end{abstract}

\section{Introduction}\label{s:introduction}

Motion segmentation algorithms are an integral part of several research fields ranging from learning motion primitives from human demonstration, action and intention recognition to rehabilitation tasks such as supervising the exercises of patients.
%
While the evaluations of algorithms in works on motion segmentation are in most cases sufficient, if considered indi\-vidually, problems arise as soon as different algorithms are to be compared to each other.
This can be affiliated to two main causes.
The first cause is the data used for the evaluation.
On the one hand, it is hard to acquire the needed data to evaluate the algorithms, which typically leads to sparse datasets.
On the other hand, labelling the data with a generally accepted ground truth is equally difficult as well as tedious, since, depending on the use case, the required ground truth can be quite different.
Some algorithms aim to segment each step of a walking person, while others merely target to distinguish between walking and not-walking persons.
Often the data used to evaluate an algorithm is not publically available, which hinders a reproducible evaluation as well.
The second cause is the variety of measures used to rate the quality of segmentation.
With different measures being used, no obvious or immediate comparability is given.
Apart from that, specific measures can be more or less suitable to reveal certain advantages or disadvantages.

Therefore, and to support the evaluation of both existing and new human motion segmentation algorithms, we propose a unifying framework, which provides programming interfaces to a set of popular programming languages.
The framework features a novel evaluation approach, termed \textit{Integrated Kernel approach}, which allows for an intuitive and semantic result interpretation in the domain of motion segmentation.
In addition, it includes established evaluation approaches known from the literature.
We also suggest a \textit{Motion Segmentation Point Hierarchy} to generate a ground truth which facilitates the evaluation of algorithms for various use cases.
The datasets and the corresponding ground truth data of human whole-body motion recordings are included as well.

\begin{figure}[t]
    \includegraphics[width=\linewidth]{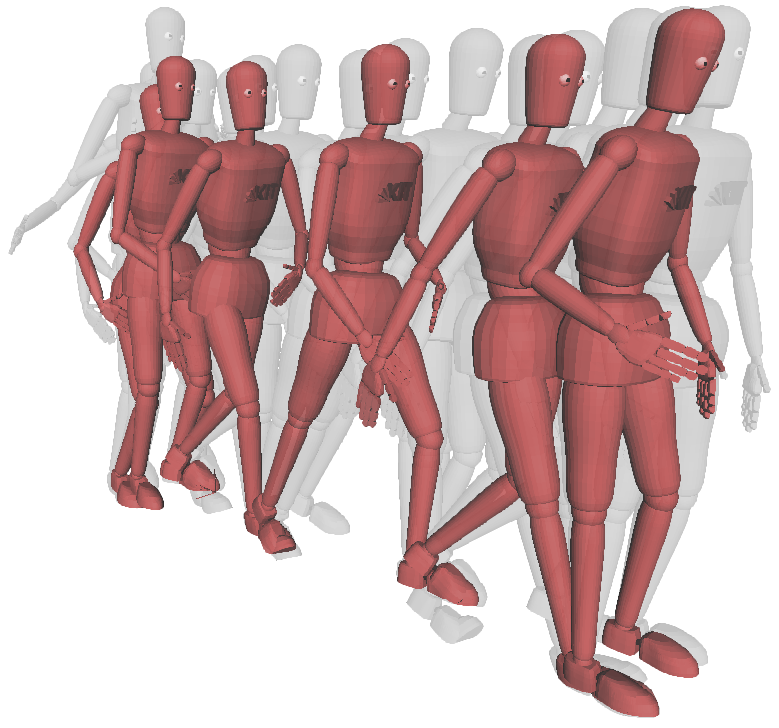}

    \caption{A motion recording of a subject jumping, then walking three steps. Individual frames are depicted grey, while key frames, which correspond to a segmentation boundary, are red.\vspace*{1.5em}}
    \label{fig:motion-recording}
\end{figure}

The rest of the paper is structured as follows:
In \refsection{s:related-work} we will give a brief overview on motion segmentation algorithms and how they were evaluated, specifically focusing on the used measures and datasets.
\refsection{s:metrics} will enumerate the various quality measures the framework uses for evaluation, including the Integrated Kernel approach we propose.
The Motion Segmentation Point Hierarchy we suggest in \refsection{s:msp-hierarchy} was used to label the provided motion recordings with a ground truth.
In \refsection{s:framework} we show the basic structure of the framework and how it utilises the concepts we propose.
After that, we will illustrate how we evaluated the framework itself in \refsection{s:evaluation}.
Finally, we will give a brief conclusion and an outlook for possible future work in \refsection{s:conclusion}.

\section{Related Work}\label{s:related-work}

\vspace*{3mm}

Early work in the field of motion segmentation was done by Fod \etal{} in \cite{fod2002}, where two solutions to the problem have been proposed.
The first one centres around Zero Velocity Crossings (ZVCs) of joint angle velocities, the other one segments whenever the Sum of Squared Angular Velocities (SSAVs) falls below a certain threshold.
The focus of the work was on data compression, therefore the evaluation concentrated on the reconstruction error, rather than on the segmentation quality. These approaches have also been used as the basis for other algorithms such as in~\cite{liebermann2004} and~\cite{lin2014b}.

Three other approaches were suggested by Barbič~\etal{} in~\cite{barbic2004} utilising Principal Component Analysis (PCA), Probabilistic PCA (PPCA) and Gaussian Mixture Models respectively.
The idea behind the PCA and PPCA approach is that, for each motion, a lower-dimensional representation can be derived.
As soon as the reconstruction error of the representation peaks, they assume that a transition into a subsequent motion occurred.
The evaluation was performed on a motion capture database consisting of $14$ motion recordings, where each motion recording contained about $10$ motions.
The measures used to evaluate the segmentation quality were the $precision$ and $recall$ measures.
For both measures, as well as for similar measures like $accuracy$, the true positives/negatives and false positives/negatives ($tp$, $tn$, $fp$, $fn$) are counted and (partially) included in the calculation of the corresponding score.
Counting the $tp$, $tn$, $fp$ and $fn$ is usually done by comparing the result of the algorithm to a ground truth of labelled data.
This proves to be problematic for motion data, because for an algorithm segmentation point, which is off by only a single frame, both an $fp$ and an $fn$ will be counted.
Depending on the frame rate of the recording, the error is negligibly small, considering that a frame usually represents only a few milliseconds.
The penalty for small errors like this should thus be reduced.
To account for this, the data was labelled with intervals rather than segmentation points.
For each segmentation point in an interval, the corresponding segmentation point of the algorithm would be counted as a $tp$, a segmentation point outside an interval as $fp$, and an interval without any segmentation points as $fn$.
Similar evaluations, using segmentation intervals rather than segmentation points, were also used in other works, for example in~\cite{lin2014a} and \cite{kulic2008}.
Obviously, the classification result heavily depends on the chosen size of the interval.

A more recent motion segmentation approach, proposed by Kulić \etal{} in~\cite{kulic2008} and~\cite{kulic2009}, is based on a general segmentation approach of time series data by Kohlmorgen and Lemm~\cite{kohlmorgen2001}, utilising Hidden Markov Models~(HMMs).
The evaluation for both works was performed on a single twenty-minute motion recording, where a single subject continuously performed $45$ different motion types.
In~\cite{kulic2009}, Kulić~\etal{} used an additional four-minute motion recording.
The motion recordings were labelled with a ground truth and afterwards compared with the actual segmentation of the algorithm.
The applied measures in the evaluation were the $recall$ and $accuracy$ measures.
To avoid the previously mentioned problems, Kulić~\etal{} allowed algorithm segmentation points to be shifted up to \SI{120}{\milli\second} (equivalent to $4$ frames for their motion recording) relative to a ground truth segmentation point.
The approach proposed in~\cite{kulic2008} has also been applied to data collected from inertial measurement units by Aoki \etal{} in~\cite{aoki2013}.
Varadarajan~\etal{} utilised HMMs to segment recordings of surgical hand mo\-tions in~\cite{varadarajan2009}.
HMMs have also been included in many other algorithms that segment or classify motions, as in \cite{lin2014b} and \cite{sanmohan2009, lee1999, bernardin2005, vicente2007}.

Lin~\etal{} suggested an algorithm using various machine learning approaches to train a binary classifier in~\cite{lin2014a}.
They applied multiple different classification methods to differentiate between segmentation points and non-segmentation points, namely k-Nearest Neighbour, Support Vector Machines, and Neural Networks.
Support Vector Machines were determined to be the most appropriate for the task.
The dataset used for the evaluation consisted of $20$ motion recordings of unique subjects performing $20$ repetitions of $5$ different rehabilitation exercises.
Before and after each ground truth segmentation point, \SI{200}{\milli\second} (equivalent to $25$ frames) were determined to be the interval where a segmentation may occur.
The measures which were utilised for the evaluation were the $F_1$-score and a novel measure termed $F_{1}^{Class}$, which additionally takes $tn$ into account.

In \cite{lin2014c}, Lin \etal{} applied the approach from \cite{lin2014a} to the same dataset that was used by Kulić~\etal{} in \cite{kulic2009}, which consists of whole-body motions.
Similar results were reported, suggesting that the approach is applicable to whole-body motions.
Classifying data points as segmentation points or non-segmentation points is a novel idea in the field of motion segmentation.
Similar approaches usually tend to classify based on motion primitives, e.g.~\cite{vicente2007, berlin2012} and \cite{zhao2013}.

W\"achter and Asfour proposed a hierarchical segmentation approach in \cite{waechter2015}, which segments based on object relations in \textls[-5]{the recording.
Hence, whenever a subject picks up or puts} down an object in the scene, a segmentation point is reported.
The result is then segmented further based on end effector trajectories.
In the evaluation of this paper, a new measure was introduced, which determines the total segmentation error compared to a ground truth.
Similarly, in \cite{mandery2015b}, Mandery~\etal{} present an approach for the segmentation of human whole-body motions based on a heuristic detection of support contacts with environmental elements.

In \cite{zhou2012}, Zhou~\etal{} used raw video data only, to obtain a hierarchy of segmentation points in return.
The problem was formulated as a minimisation of hierarchically aligned cluster analysis.
A downside of this approach is that the number of segments needs to be known in advance.
The idea of hierarchically splitting motions was also presented in other works, for example in \cite{sanmohan2009, waechter2015, kulic2010} and \cite{voegele2014}.

\textls[-10]{There are many more algorithms that adapt entirely different} approaches, but also segment motion recordings in~\cite{lin2016a, mueller2006, lu2004, lan2015, krueger2017}.

As also mentioned by Lin \etal{} in \cite{lin2016b}, where a broad overview on motion segmentation algorithms can be found, comparing algorithms is challenging due to the variety of measures that were used or the absence of actual segmentation quality scores in the evaluation.

\section{Segmentation Evaluation Measures}\label{s:metrics}

The measures which will be applied to evaluate the segmentation quality are $precision$, $recall$, $accuracy$, $F_1$-score, $F_1^{Class}$-score (proposed in \cite{lin2014a}) and the Matthews Correlation Coefficient ($\mcc$, proposed in \cite{matthews1975}).
They are defined as follows:
\begin{align*}
precision &= \frac{tp}{tp + fp}, \\
recall &= \frac{tp}{tp + fn}, \\
accuracy &= \frac{tp + tn}{tp + tn + fp + fn}, \\
F_{1} &= \frac{2 \cdot tp}{2 \cdot tp + fp + fn}, \\
F_{1}^{Class} &= \frac{2 \cdot (tp + tn)}{2 \cdot (tp + tn) + fp + fn}, \\
\mcc &= \frac{tp \cdot tn - fp \cdot fn}{\sqrt{(tp+fp)(tp+fn)(tn+fp)(tn+fn)}}.
\end{align*}

Apart from those measures, the Penalised Squared Minimum Error measure ($\psme$, suggested in \cite{waechter2015}) will be applied as well, which is defined as
$$ \psme = p \cdot (fp + fn) + \sum_i \min_j((k_{s,i} - k_{g,j})^2), $$
where $p$ is a penalty factor, $k_{s,i}$ the $i$th algorithm segmentation point and $k_{g,j}$ the $j$th ground truth segmentation point.
It differs from the other measures in that it is an absolute error measure rather than a ratio.
Therefore, this measure will be treated separately in the following.

As already mentioned before, the conventional approach to determine the $tp$, $tn$, $fp$ and $fn$ does not provide meaningful results in the domain of motion segmentation.
Motion recordings usually contain several thousand frames, whereas the number of segmentation points tends to be much smaller.
A segmentation point which is only a few milliseconds off relative to the ground truth will be counted as both $fp$ and $fn$ instead of a $tp$, even though results with such negligible errors would be highly desirable.

To overcome this, many works allowed algorithm segmentation points to be in the proximity of a ground truth segmentation point and still be counted as $tp$, for example~\cite{barbic2004, kulic2008, kulic2009, lin2014a} and \cite{lin2014c}.
We termed this method the \textit{Margin approach} and defined it similar to \cite{kulic2009}.
A segmentation point $x$ lies in the margin of another segmentation point $y$, if the temporal distance between $x$ and $y$ is smaller than the margin size, which we set to \SI{200}{\milli\second} (equivalent to $20$ frames for the provided motion recordings).
An algorithm segmentation point is counted as a $tp$, if it lies in the margin of a ground truth segmentation point.
If there are several algorithm segmentation points in the margin of a ground truth segmentation point, one will be counted as a $tp$, while the others are counted as $fp$.
If there is an algorithm segmentation point which is not in the margin of any ground truth segmentation point, it will be counted as an $fp$ as well.
An $fn$ will be counted, if there is no algorithm segmentation point within the margin of a ground truth segmentation point.
Eventually, the number of $tn$ can be derived with $tn = f_{max} - tp - fp - fn$, where $f_{max}$ is the number of frames.
While measures evaluated with $tp$, $tn$, $fp$ and $fn$ determined by the Margin approach provide more meaningful results, there are still problems to consider.
First of all, the Margin approach does not take temporal distances into account.
The scores of the measures for a perfect segmentation would not differ from the scores where a perfect measure was shifted to one side by the size of the margin.
Another problem is that the Margin approach may count a $tp$ twice when the margins of two algorithm segmentation points overlap and the ground truth segmentation point lies in the intersection.
The opposite case also applies, where two ground truth segmentation points overlap and an algorithm segmentation point lies in the intersection.

We therefore suggest a novel approach to determine the total number of $tp$, $tn$, $fp$ and $fn$, which makes use of a non-negative kernel function $k(x)$ such that $ \int{k(x) dx = 1} $.
Hence we call it the \textit{Integrated Kernel approach} (InK approach).
We chose the kernel function $k$ to be the Gaussian distribution.
Contrary to the Margin approach, the InK approach takes temporal distances into account.
The basic principle is to combine two functions, one depicting the ground truth seg\-mentation points ($f_{gt}$) and one the algorithm segmentation points ($f_{s}$), and interpret the integrals of the results as $tp$, $fp$ and $fn$.
This means that we allow $tp$, $fp$, $fn$ and $tn$, which can be derived by the former three, to be real numbers instead of integers.
We define a segmentation point to be the time stamp, where one segment ends and another one starts.
Let $GT \subset \mathbb{R}$ be the set of all segmentation points in the ground truth and $S \subset \mathbb{R}$ the set of all segmentation points reported by an algorithm.
We then define the ground truth function $f_{gt}$ as
$$ f_{gt}(x) = - \sum_{f \in GT} k(x-f). $$
Similarly, we define the segmentation function $f_{s}$ as
$$ f_{s}(x) = \sum_{f \in S} k(x-f). $$
The combined error function $e_{c}$ is now defined as follows:
$$ e_{c} = f_{s} + f_{gt}. $$

\begin{figure}[ht]
    \includegraphics[width=\linewidth]{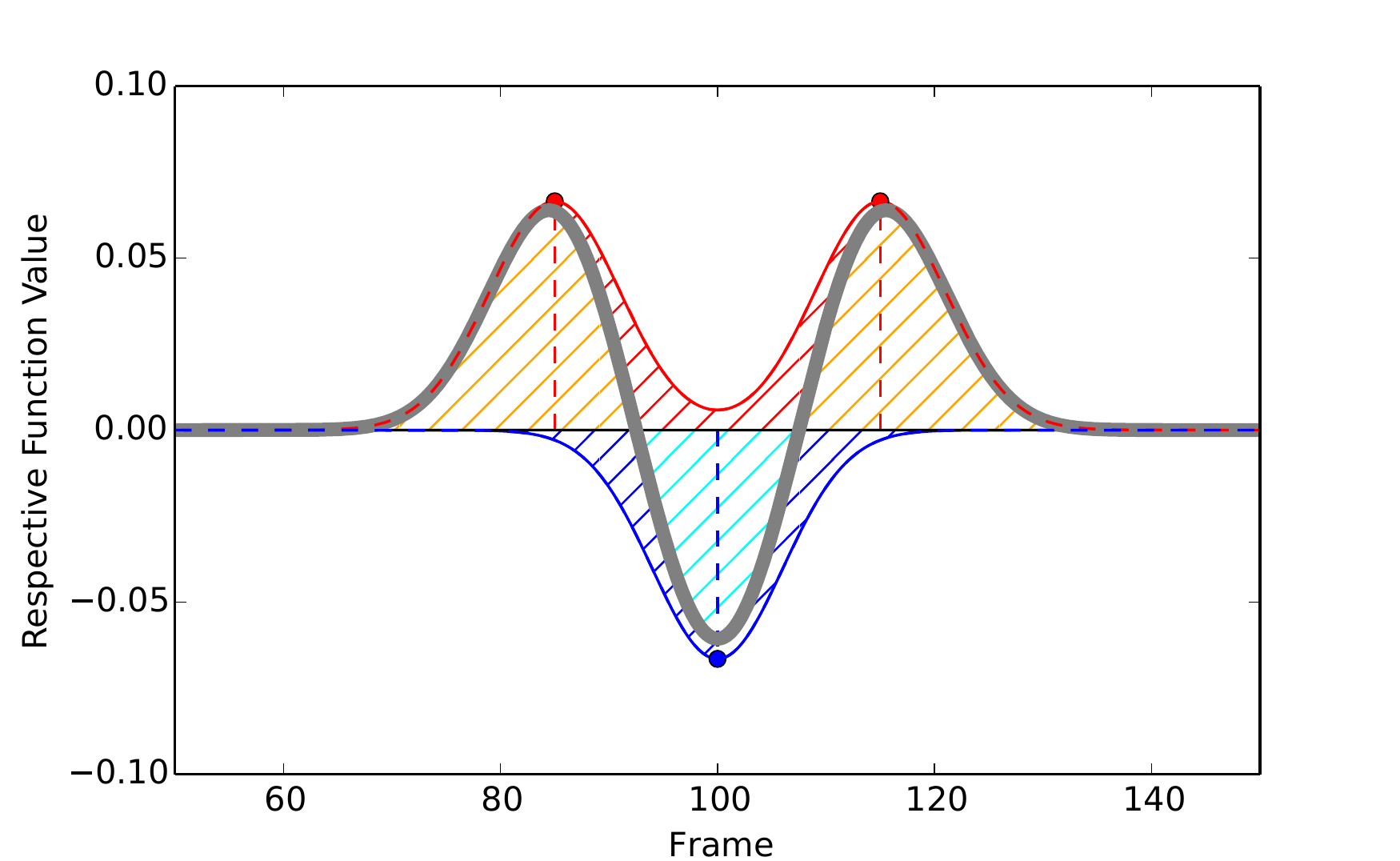}

    \caption{Example of the Integrated Kernel approach with two segmentation points of an algorithm (red dots) and a ground truth of one segmentation point (blue dot). Red function: $f_{s}$; Blue function: $f_{gt}$; Grey function: $e_{c}$; Orange area: $A_{fp}$; Cyan area: $-A_{fn}$; Red area: $A_{tp}$.}
    \label{fig:kernel-metric-intro}
\end{figure}

The functions $e_{c}$, $f_{s}$ and $f_{gt}$ are plotted exemplarily in \autoref{fig:kernel-metric-intro}, where an algorithm detected two segmentation points (red dots) while the ground truth only shows one segmentation point (blue dot).
The red function denotes $f_{s}$ and the blue function $f_{gt}$.
Adding these functions results in the function $e_{c}$ outlined in grey.

The combined error function $e_{c}$ can now be split into two distinct error functions depending on the sign, namely
$$ e_{fp}(x) = \begin{cases} e_{c}(x) &\mbox{if } e_{c}(x) > 0 \\
0 & \mbox{otherwise} \end{cases} $$
as the $fp$ error function, and
$$ e_{fn}(x) = \begin{cases} e_{c}(x) &\mbox{if } e_{c}(x) < 0 \\
0 & \mbox{otherwise} \end{cases} $$
as the $fn$ error function.
We now define the false positive area $A_{fp}$ as
$$ A_{fp} = \int e_{fp}(x) dx $$
and likewise the false negative area $A_{fn}$ as
$$ A_{fn} = -\int e_{fn}(x) dx. $$

In \autoref{fig:kernel-metric-intro} $A_{fp}$ is depicted with the orange area and $A_{fn}$ with the cyan-coloured area respectively (with negative sign).

We can now proceed to calculate the true positive area $A_{tp}$ as follows:
$$ A_{tp} = \int f_{s}(x) dx - A_{fp} = \abs{S} - A_{fp}, $$
which is depicted by the red area in \autoref{fig:kernel-metric-intro}.
Given the number of total frames $f_{max}$ we can now also calculate the true negative area $A_{tn}$ as follows:
$$ A_{tn} = f_{max} - A_{tp} - A_{fp} - A_{fn}. $$

\begin{figure}[ht]
    \includegraphics[width=\linewidth]{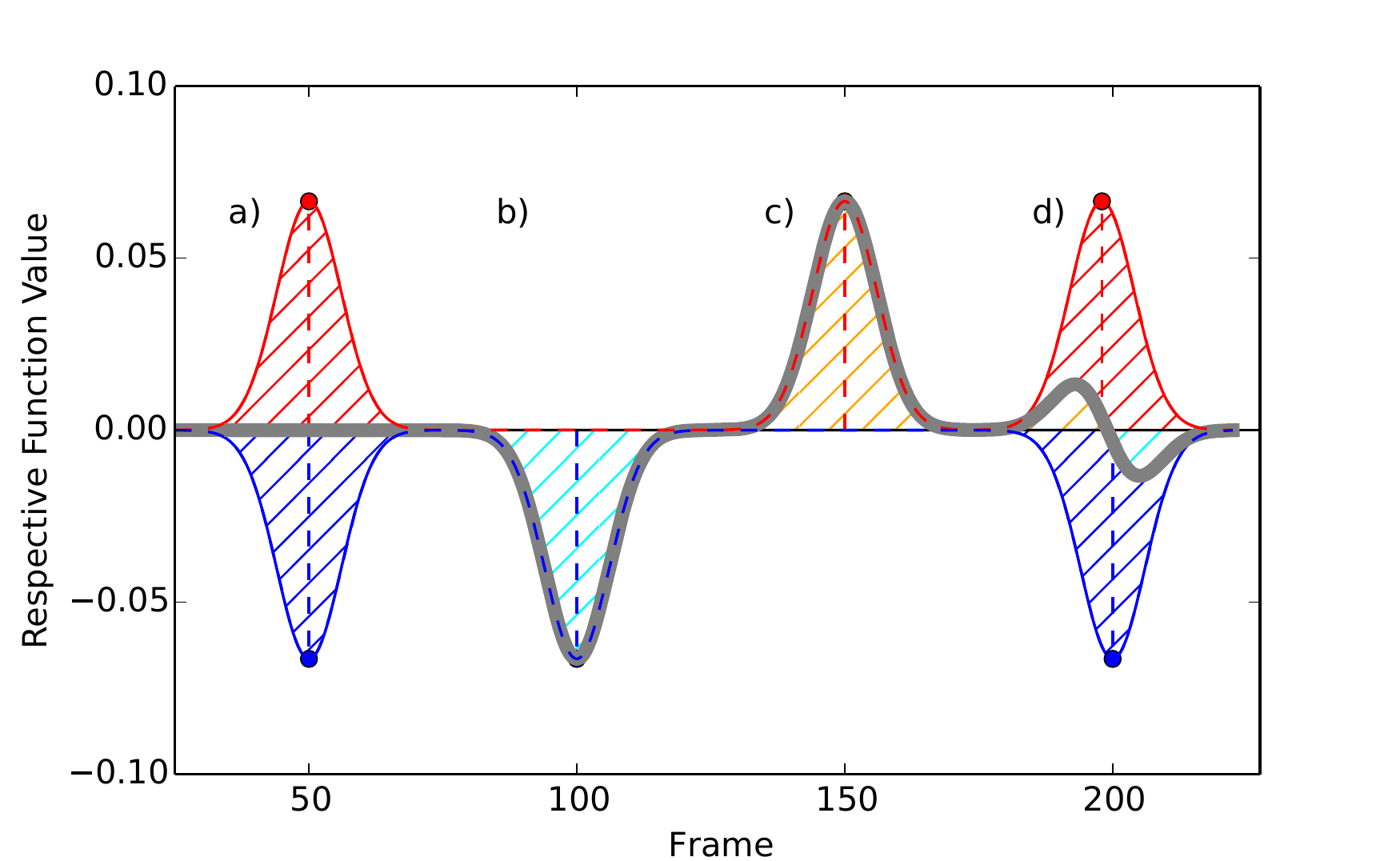}

    \caption{The four trivial cases of the Integrated Kernel approach. a) Perfect segmentation; b) False negative; c) False positive; d) Close segmentation. The colours of the plotted functions and areas are defined as in \autoref{fig:kernel-metric-intro}.}
    \label{fig:kernel-metric}
\end{figure}

We identified four different trivial cases which can occur using this approach.
They are shown in \autoref{fig:kernel-metric}.

The first case, a), is a perfect segmentation, resulting in no error since both $f_{s}$ and $f_{gt}$ annihilate each other in $e_{c}$ for the corresponding frames.
In case b) and c) an $fn$ (cyan) and an $fp$ (orange) can be observed, respectively.
In both cases, our approach does not differ from the known definition because the integral of the kernel function was defined to equal $1$.
The last case, d), is the most interesting case, where the segmentation is not perfect, yet $f_{s}$ and $f_{gt}$ overlap because they are very close to each other.
The algorithm segmented earlier than it was denoted in the corresponding ground truth, which results in a partial annihilation of $f_{s}$ and $f_{gt}$.
The resulting orange and cyan areas are counted as partial $fp$ and $fn$ respectively, while the red area will be counted as a partial $tp$.

To make the Margin approach comparable with the InK approach, we set $\sigma$ of the Gaussian distribution to \SI{66.67}{\milli\second}.
This way, roughly \SI{99.7}{\percent} of the area of the Gaussian distribution's integral will be in the range of $\pm 3 \sigma \approx \SI{200}{\milli\second}$, which in turn corresponds to the margin size of \SI{200}{\milli\second} (equivalent to $20$ frames for the provided motion recordings).

It is interesting to note that other kernel functions can be used as well.
The conventional approach, for example, can be mimicked if the kernel function is set to the Dirac delta function ($\delta$ function).



\vspace{1em}

\section{Motion Segmentation Point Hierarchy}\label{s:msp-hierarchy}

As already mentioned before, one major problem with evaluations of works on motion segmentation is that there is no generally applicable ground truth, because segmentation algorithms are usually designed and fine-tuned for specific use cases.
For example, one algorithm may be designed as an activity recognition component in a smart home environment, \textls[-5]{while another algorithm may be used on a robot which is} capable of learning new motions by demonstration of a human teacher.
Each of these use cases come with certain \textit{granularities} for which the given algorithm performs best -- e.g. the smart home algorithm would not be able to distinguish motion primitives as well as the one used on a robot that depends on it in order to learn new movements.

\begin{figure}[ht]
    \includegraphics[width=\linewidth]{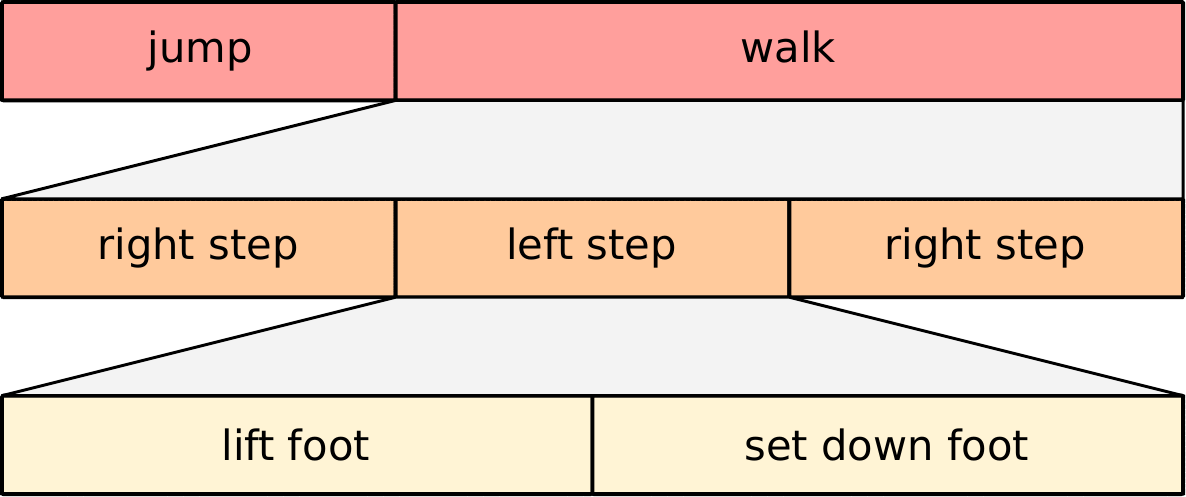}
    \caption{Depiction of the Motion Segmentation Point Hierarchy on an exemplary motion recording of a person jumping, then walking three steps as shown in \autoref{fig:motion-recording}.
    Red/top: Segments of rough granularity.
    Orange/middle: Segments of medium granularity for the \textit{walk}-segment.
    Yellow/bottom: Seg\-ments of fine granularity for a single \textit{step}-segment.}
    \label{fig:msp-hierarchy}
\end{figure}

We therefore propose a Motion Segmentation Point Hierarchy, in which each ground truth segmentation point is tagged with a granularity.
In total there are three granularities and they are defined as follows:
\begin{itemize}
    \item{\textit{Rough granularity}: Used to distinguish activities.\\ Example: Jumping vs. walking}
    \item{\textit{Medium granularity}: Used to distinguish actions.\\ Example: Stepping with left vs. right foot}
    \item{\textit{Fine granularity}: Used to distinguish motion primitives.\\ Example: Lift foot for step vs. set down foot}
\end{itemize}
A motion recording labelled with these tagged segmentation points could look as depicted in \autoref{fig:msp-hierarchy}.

We make use of the granularities while evaluating the segmentation quality of an algorithm.
For each granularity, the evaluation process will be performed once, starting with only the ground truth segmentation points tagged as rough.
After that, the ground truth segmentation points tagged as medium will be added and the evaluation process will be repeated.
In the last stage, the remaining ground truth segmentation points, tagged as fine, will be added and again, the evaluation process will be performed.
Therefore, for each measure evaluating the segmentation quality, we receive three scores, one for each granularity.

While the introduction of different granularities increases the complexity of labelling the ground truth, the Motion Segmentation Point Hierarchy serves as a guideline for annotators, making the labelling process more intuitive after a brief instruction.

\section{Motion Segmentation Evaluation Framework}\label{s:framework}

We propose a framework which is able to run and evaluate motion segmentation algorithms.
It is divided into several modules, features an open source license, and is available at \url{https://gitlab.com/h2t/kit-mseg}.

The first module is the \textit{core module}, which provides a graphical user interface, several evaluation measures, datasets provided with a ground truth, and an API to communicate with the motion segmentation algorithms.
It is integrated into the ArmarX framework introduced in \cite{vahrenkamp2015}.

The remaining modules are implemented in the programming languages most commonly used for motion segmentation algorithms, which we identified to be \textsc{Matlab} and \cpp{}, as well as Python and Java.
They serve as \textit{Programming Language Interfaces} (PLIs), which abstract the communication to the core module over the network.
They offer an API to the programming languages the algorithms were implemented in, easing the integration into the framework without knowledge of its implementation details.
The PLIs are mandatory to ensure the communication between the core module and the algorithms, even though we will not always mention them explicitly in the following.

The communication between the core module and the algorithms takes place in two directions.
On the one hand, the core module can set algorithm parameters and may order the algorithm to segment a user-selected dataset.
The algorithm, on the other hand, may request data to segment or data provided with a ground truth to train (if it is learning-capable), it can notify the core module about available parameters for the user to tweak in the graphical user interface, and eventually it can report segmentation points to the core module.

After an algorithm was ordered to segment a motion recording, the core module starts the evaluation as soon as the segmentation results are reported.
The general evaluation process includes the calculation of $tp$, $tn$, $fp$ and $fn$ employing the Margin approach and $A_{tp}$, $A_{tn}$, $A_{fp}$ and $A_{fn}$ employing the InK approach.
Using these numbers, the measures listed in \refsection{s:metrics} are computed for each approach.
The general evaluation process is executed three times, once for each granularity, and from rough over medium to fine.
If the algorithm was ordered to segment a dataset rather than a single motion recording, the described process is repeated for each motion recording.
Finally, the evaluation results are grouped and displayed in the graphical user interface.

Depending on whether a motion segmentation algorithm requires training data, the dataset is split into five parts of roughly equal size and a $5$-fold cross validation is performed.
However, in order to make the evaluation results reproducible, the individual parts are not randomly assigned but statically split.

The evaluation results shown in the graphical user interface also include general information, e.g. whether the algorithm fetched the motion recordings frame-by-frame or at once, whether the algorithm used the API to receive training data or not, or what dataset was used for the evaluation.
To ensure that evaluation results maintain comparability in the future and remain reproducible, relevant information like the parameter configuration, core module version, and dataset version are included as well.




\section{Evaluation}\label{s:evaluation}

In the following subsections, both the InK approach and the evaluation framework will be evaluated.

\subsection{Integrated Kernel Approach}

\begin{figure}[ht]
    \includegraphics[width=\linewidth]{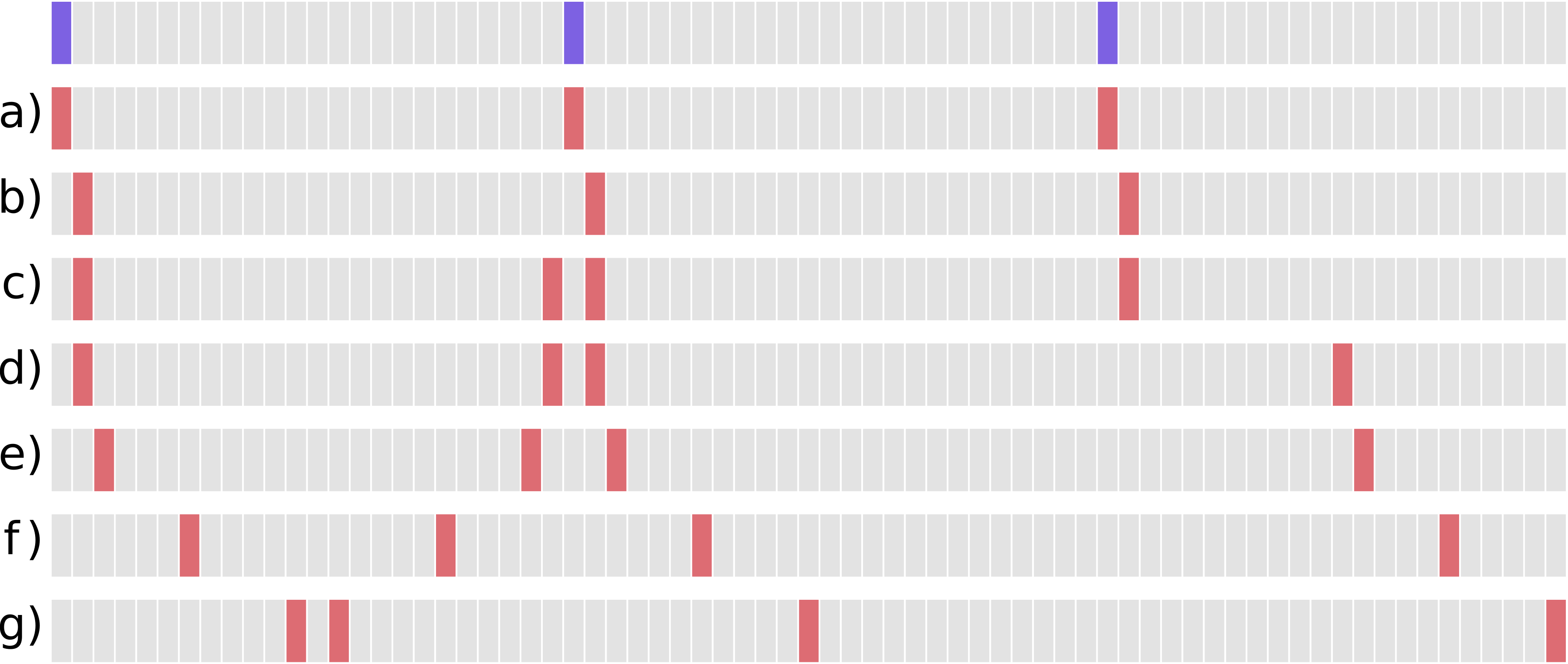}
    \caption{Data used to exemplarily evaluate the Integrated Kernel approach. Top row (blue) is the ground truth for reference; a) is a perfect segmentation; b) is $1$ frame off from the ground truth; c) introduces an additional segmenta\-tion point; d) is additionally $10$ frames off for the last key frame; e), f) and g) increase the error of their respective successor by $1$, $4$ and $5$ frame(s).}
    \label{fig:ikm-eva}
\end{figure}

Initially, the InK approach was evaluated without an actual segmentation algorithm using synthetic data.
Seven varying segmentation scenarios were therefore defined as shown in \autoref{fig:ikm-eva}, where the segmentation quality ranges from perfect to very poor.

\textls[7]{For comparison, these scenarios were evaluated with the} conventional $F_1$-score, the $F_1$-score using the Margin approach with a margin size of $5$ frames, and the $F_1$ score using the InK approach with a standard deviation of $\sigma=1.67$.
The results are listed in \autoref{tab:ikm-eva-res}.

\begin{table}[ht]
    \caption{Comparison between the scores of $F_1$, $F_1$ using the Margin approach with a margin size of $5$ ($F_1^{M}$) and $F_1$ using the InK approach with $\sigma=1.67$ ($F_1^{InK}$) for seven exemplary segmentation scenarios as shown in \autoref{fig:ikm-eva}.}
    \centering
    \begin{tabular}{ l r r r }
        \toprule[1pt]
        Scenario & $F_1$ & $F_1^{M}$ & $F_1^{InK}$ \\
        \midrule[0.5pt]
        a) & $1.00$ & $1.00$ & $1.00$ \\
        b) & $0.00$ & $1.00$ & $0.76$ \\
        c) & $0.00$ & $0.86$ & $0.72$ \\
        d) & $0.00$ & $0.57$ & $0.50$ \\
        e) & $0.00$ & $0.57$ & $0.44$ \\
        f) & $0.00$ & $0.00$ & $0.06$ \\
        g) & $0.00$ & $0.00$ & $0.00$ \\
        \bottomrule[1pt]
    \end{tabular}
    \label{tab:ikm-eva-res}
\end{table}

The conventional $F_1$ score, as described previously, does not regard the offset of one frame at all, counting these points as both an $fp$ and an $fn$, which leads to a score of $0.00$ beginning from scenario b).
The approach using margins, however, counts them as $tp$ as long as they lie inside the margin.
Hence the score is still $1.00$ for scenario b) and the scores for scenarios d) and e) are equal despite the segmentation in scenario d) being better.
In scenario f), the segmentation points shift out of the margin, resulting in a score of $0.00$ immediately.
The InK approach provides considerably more continuous scores, depending on the temporal distances that the individual segmentation points are off relative to the ground truth.
For all measures, the score will be $1.00$ for a perfect segmentation.
However, the Margin approach also reports a score of $1.00$ if a perfect segmentation was shifted to one side by the size of the margin, which does not reflect the error at all.

\subsection{Framework}

The framework was evaluated by integrating motion seg\-mentation algorithms from the literature.
The main focus was to gather various types of algorithms (online, offline, learning-capable) to evaluate the core module on the one hand, and to have them implemented in \cpp{}, Python, Java and \textsc{Matlab} to evaluate the PLIs on the other hand.

\textls[17]{We already had access to the source code of a re}-implementation of the ZVC approach by Fod \etal{}, implemented in \cpp{}.
Since we had no access to an algorithm implemented in Python or Java, we decided to implement further algorithms.
Firstly, the SSAV approach by Fod \etal{} was implemented in Python and modified to be more appropriate for whole-body motion.
While the original SSAV approach segments only if the sum of squared angular velocities falls below a threshold $t$, the modified version seeks for a minimum within a sliding time window.
For a segmentation point to be reported, this minimum has to be below the threshold $t$, but still above another threshold, which filters noise.
Secondly, we implemented the PCA approach by Barbič \etal{} in Java.

To gather data, the KIT Whole-Body Human Motion Database, introduced in \cite{mandery2015} and \cite{mandery2016}, was browsed for suitable motion recordings.
The criteria were that the recordings are relatively long and feature at least two different activities.
The selected motion recordings were labelled using the Motion Segmentation Point Hierarchy.

The evaluation was executed on a sample motion recording (\texttt{martial-arts\_01\_v1.xml}) of a subject performing martial art movements.
For both the Margin approach and the InK approach, and for each of the three granularities, the measures $precision$, $recall$, $accuracy$, $F_1$-score, $F_1^{Class}$-score and $\mcc$ were calculated once.
The results of the evaluation are shown in \autoref{tab:eva-zvc}, \autoref{tab:eva-ssav} and \autoref{tab:eva-pca}.

The results show that the $accuracy$ ranges from $0.94$ to $1.00$, whereas the $F_1^{Class}$-score never undercuts $0.97$ and also peaks to $1.00$ in some cases.
This can be affiliated to the large number of $tn$ compared to $tp$, which are also taken into account for these measures.
Having such a small range in either case does not give much information about the actual segmentation quality.

Both $precision$ and $recall$ are much more descriptive and reveal how the algorithm behaves.
For example, the low $precision$ for the ZVC approach at the rough granularity in \autoref{tab:eva-zvc} shows that the algorithm produced a lot of $fp$, because there is only a small number of segmentation points in the ground truth for a rough granularity.
The ZVC approach, however, tends to oversegment.
This is reflected by the much higher $precision$ considering a fine granularity.

In contrast to the ZVC approach, the PCA approach from Barbič requires a certain amount of time to form a lower-dimensional representation of the motion data.
While computing the representation, the algorithm is not able to detect segmentation points, which leads to much fewer segments.
\autoref{tab:eva-pca} shows that the $recall$ of the algorithm for the InK approach worsens increasingly with each finer granularity level.
This can be attributed to an increasing number of $fn$.
The increasing $recall$ for the Margin approach at a fine granularity is an artefact and will be discussed later.

Comparing the Margin approach with the InK approach, it can be seen that the latter changes steadily in accordance with the temporal distance.
For example, the $precision$ of the PCA approach in \autoref{tab:eva-pca} for the Margin approach peaks to $1.00$ beginning at medium granularity.
The InK approach, however, reveals that the $precision$ steadily increases for each finer granularity level.

While the PCA approach returned $7$ segmentation points for the given motion recording, the data in \autoref{tab:eva-pca} shows that the Margin approach counted a total of $11$ $tp$ for the fine granularity.
An effect resulting from this can be observed in the $recall$ of the PCA approach.
The $recall$ is $0.22$ for the rough granularity, falling to $0.13$ for the medium granularity, and rising again to $0.14$ for the fine granularity as the number of $tp$ rises.
The cause for this is that, as already mentioned, the Margin approach may count a $tp$ twice when the margins of two ground truth segmentation points overlap and an algorithm segmentation point lies in the intersection of them.
On the contrary, the InK approach is not prone to fluctuations like this, as it shows a steady decrease of the $recall$ from $0.22$ over $0.07$ to $0.06$ respectively.

\vspace{2mm}

\begin{table}[!h]
    \centering
    \caption{Evaluation results for the ZVC approach by Fod~\cite{fod2002}. Scores of the Margin approach compared to those of the Integrated Kernel approach (InK), further divided by granularity.}
    \begin{tabular}{ l r r r r r r }
        \toprule[1pt]
        & \multicolumn{3}{c}{\bf Margin} & \multicolumn{3}{c}{\bf InK} \\
        \cmidrule{2-4} \cmidrule(l{2pt}){5-7}
        \scriptsize{Granularity} & \scriptsize{Rough} & \scriptsize{Med.} & \scriptsize{Fine} & \scriptsize{Rough} & \scriptsize{Med.} & \scriptsize{Fine} \\
        \midrule[0.5pt]
        $tp$/$A_{tp}$ &    $9$ &   $52$ &   $80$ &    $8.3$ &   $47.7$ &   $68.4$ \\
        $tn$/$A_{tn}$ & $2600$ & $2597$ & $2538$ & $2598.3$ & $2593.7$ & $2586.4$ \\
        $fp$/$A_{fp}$ &  $174$ &  $133$ &  $164$ &  $175.7$ &  $136.3$ &  $115.6$ \\
        $fn$/$A_{fn}$ &    $0$ &    $1$ &    $1$ &    $0.7$ &    $5.3$ &   $12.6$ \\
        \midrule[0.5pt]
        $precision$   & $0.05$ & $0.28$ & $0.33$ &   $0.05$ &   $0.26$ &   $0.37$ \\
        $recall$      & $1.00$ & $0.98$ & $0.99$ &   $0.92$ &   $0.90$ &   $0.84$ \\
        $accuracy$    & $0.94$ & $0.95$ & $0.94$ &   $0.94$ &   $0.95$ &   $0.95$ \\
        $F_{1}$       & $0.09$ & $0.44$ & $0.49$ &   $0.09$ &   $0.40$ &   $0.52$ \\
        $F_1^{Class}$ & $0.97$ & $0.98$ & $0.97$ &   $0.97$ &   $0.97$ &   $0.98$ \\
        $\mcc$        & $0.21$ & $0.51$ & $0.55$ &   $0.20$ &   $0.47$ &   $0.54$ \\
        \bottomrule[1pt]
    \end{tabular}
    \label{tab:eva-zvc}
\end{table}

\begin{table}[!h]
    \centering
    \caption{Evaluation results for the SSAV approach by Fod~\cite{fod2002}. Scores of the Margin approach compared to those of the Integrated Kernel approach (InK), further divided by granularity.}
    \begin{tabular}{ l r r r r r r }
        \toprule[1pt]
        & \multicolumn{3}{c}{\bf Margin} & \multicolumn{3}{c}{\bf InK} \\
        \cmidrule{2-4} \cmidrule(l{2pt}){5-7}
        \scriptsize{Granularity} & \scriptsize{Rough} & \scriptsize{Med.} & \scriptsize{Fine} & \scriptsize{Rough} & \scriptsize{Med.} & \scriptsize{Fine} \\
        \midrule[0.5pt]
        $tp$/$A_{tp}$ &    $6$ &   $33$ &   $36$ &    $4.9$ &   $29.3$ &   $31.5$ \\
        $tn$/$A_{tn}$ & $2705$ & $2690$ & $2662$ & $2703.9$ & $2684.3$ & $2658.5$ \\
        $fp$/$A_{fp}$ &   $69$ &   $40$ &   $40$ &   $70.1$ &   $45.7$ &   $43.5$ \\
        $fn$/$A_{fn}$ &    $3$ &   $20$ &   $45$ &    $4.1$ &   $23.7$ &   $49.5$ \\
        \midrule[0.5pt]
        $precision$   & $0.08$ & $0.45$ & $0.47$ &   $0.07$ &   $0.39$ &   $0.42$ \\
        $recall$      & $0.67$ & $0.62$ & $0.44$ &   $0.55$ &   $0.55$ &   $0.39$ \\
        $accuracy$    & $0.97$ & $0.98$ & $0.97$ &   $0.97$ &   $0.98$ &   $0.97$ \\
        $F_{1}$       & $0.14$ & $0.52$ & $0.46$ &   $0.12$ &   $0.46$ &   $0.40$ \\
        $F_1^{Class}$ & $0.99$ & $0.99$ & $0.98$ &   $0.99$ &   $0.99$ &   $0.98$ \\
        $\mcc$        & $0.23$ & $0.52$ & $0.44$ &   $0.18$ &   $0.45$ &   $0.39$ \\
        \bottomrule[1pt]
    \end{tabular}
    \label{tab:eva-ssav}
\end{table}

\begin{table}[!h]
    \centering
    \caption{Evaluation results for the PCA approach by Barbič~\cite{barbic2004}. Scores of the Margin approach compared to those of the Integrated Kernel approach (InK), further divided by granularity.}
    \begin{tabular}{ l r r r r r r }
        \toprule[1pt]
        & \multicolumn{3}{c}{\bf Margin} & \multicolumn{3}{c}{\bf InK} \\
        \cmidrule{2-4} \cmidrule(l{2pt}){5-7}
        \scriptsize{Granularity} & \scriptsize{Rough} & \scriptsize{Med.} & \scriptsize{Fine} & \scriptsize{Rough} & \scriptsize{Med.} & \scriptsize{Fine} \\
        \midrule[0.5pt]
        $tp$/$A_{tp}$ &    $2$ &    $7$ &   $11$ &    $2.0$ &    $3.8$ &    $4.9$ \\
        $tn$/$A_{tn}$ & $2769$ & $2730$ & $2702$ & $2769.0$ & $2726.8$ & $2699.9$ \\
        $fp$/$A_{fp}$ &    $5$ &    $0$ &    $0$ &    $5.0$ &    $3.2$ &    $2.1$ \\
        $fn$/$A_{fn}$ &    $7$ &   $46$ &   $70$ &    $7.0$ &   $49.2$ &   $76.1$ \\
        \midrule[0.5pt]
        $precision$   & $0.29$ & $1.00$ & $1.00$ &   $0.28$ &   $0.55$ &   $0.71$ \\
        $recall$      & $0.22$ & $0.13$ & $0.14$ &   $0.22$ &   $0.07$ &   $0.06$ \\
        $accuracy$    & $1.00$ & $0.98$ & $0.97$ &   $1.00$ &   $0.98$ &   $0.97$ \\
        $F_{1}$       & $0.25$ & $0.23$ & $0.24$ &   $0.25$ &   $0.13$ &   $0.11$ \\
        $F_1^{Class}$ & $1.00$ & $0.99$ & $0.99$ &   $1.00$ &   $0.99$ &   $0.99$ \\
        $\mcc$        & $0.25$ & $0.36$ & $0.36$ &   $0.25$ &   $0.19$ &   $0.20$ \\
        \bottomrule[1pt]
    \end{tabular}
    \label{tab:eva-pca}
\end{table}

While $precision$ and $recall$ provide an in-depth look of an algorithm's behaviour, the $F_1$-score and $\mcc$ give a general overview.
As can be seen in the results, both measures approximately provide similar ratings.

Overall, the scores for the InK approach are lower than the corresponding ones using the Margin approach, because the InK approach takes temporal distances into account.

Running the evaluation for three granularities reveals that the algorithms considered here are best suited for quite different use cases.
The ZVC approach achieves the best scores when evaluated with a fine granularity ground truth, while the PCA approach achieves its highest scores using a rough granularity ground truth.
Due to its modifications, the SSAV approach obtains its best result for a medium granularity considering its $F_1$-score.

The $\psme$ measure was applied independently from the Margin or InK approach, since it is an absolute error measure.
The results using the same data are listed in \autoref{tab:eva-psme}.
While the score of this measure does not expose specific properties of the algorithm, it is a useful quantity to minimise when the algorithm should be optimised by tweaking its parameters.

\begin{table}[ht]
    \caption{Comparison of the scores of the $\psme$ measure for the ZVC and the SSAV approach by Fod~\cite{fod2002} and for the PCA approach by Barbič~\cite{barbic2004} considering different granularities.}
    \centering
    \begin{tabular}{ l r r r }
        \toprule[1pt]
        Algorithm & Rough & Medium & Fine \\
        \midrule[0.5pt]
        ZVC  & $1337$ & $1617$ & $1799$ \\
        SSAV &  $574$ &  $798$ &  $994$ \\
        PCA  &   $98$ &  $406$ &  $602$ \\
        \bottomrule[1pt]
    \end{tabular}
    \label{tab:eva-psme}
\end{table}

\section{Conclusion and Future Work}\label{s:conclusion}

In this paper, a framework was presented, which is able to evaluate motion segmentation algorithms on a unified testing ground, consisting of a wide range of measures and labelled datasets.
Important contributions of the work are a novel measure utilising a kernel function to ease the interpretation of scores that rate the segmentation quality of an algorithm, as well as a hierarchy for segmentation points to support several types of motion segmentation algorithms.
Motion segmentation algorithms can be evaluated regardless of whether they are implemented in \cpp{}, Java, Python or \textsc{Matlab}.

For the future, we intend to integrate further motion seg\-men\-tation algorithms to thoroughly evaluate and compare them with each other and with the already integrated ones.
For this purpose, larger datasets would be of great importance, especially in respect of motion segmentation algorithms requiring training data.
We therefore plan to gather more labelled data using the KIT Whole-Body Human Motion Database and a crowdsourcing approach similar to our previous work described in \cite{plappert2016}.


\section{Acknowledgements}
We would like to thank Dana Kulić and Jonathan Lin for the helpful discussions in the early stages of this work.

\IEEEtriggeratref{17}

\bibliographystyle{IEEEtran}
\bibliography{references}

\end{document}